\documentclass[11pt]{article}

\usepackage[preprint]{acl}
\usepackage{times}
\usepackage{latexsym}
\usepackage[T1]{fontenc}
\usepackage[utf8]{inputenc}
\usepackage{microtype}
\usepackage{inconsolata}
\usepackage{graphicx}
\usepackage{booktabs}
\usepackage{amsmath}
\usepackage{amssymb}
\usepackage{xcolor}
\usepackage{longtable}
\usepackage{url}
\usepackage{placeins}
\usepackage{enumitem}

\setlist{nosep,leftmargin=*}
\AtBeginDocument{%
  \setlength{\abovedisplayskip}{4pt plus 1pt minus 1pt}%
  \setlength{\belowdisplayskip}{4pt plus 1pt minus 1pt}%
  \setlength{\abovedisplayshortskip}{2pt plus 1pt}%
  \setlength{\belowdisplayshortskip}{2pt plus 1pt}%
}

\newcommand{\TaskCount}{40}
\newcommand{\ModelCount}{34}
\newcommand{\OpenModelCount}{25}
\newcommand{\ControlModelCount}{5}
\newcommand{\FrontierModelCount}{4}
\newcommand{\RequestCount}{1360}
\newcommand{\SpecificationPassCount}{32}
\newcommand{\SpecificationPassRate}{2.35\%}
\newcommand{\NearPassCount}{14}
\newcommand{\NearPassRate}{1.03\%}

\newcommand{\RepairPassRate}{3.24\%}

\newcommand{\CleanPassRate}{19.85\%}

\newcommand{\SourceCleanPassRate}{19.78\%}
\newcommand{\OverallRepairProgress}{27.5\%}
\newcommand{\ValidRepairProgress}{50.9\%}
\newcommand{\OverallValidUCR}{2.4\%}

\newcommand{\TargetMatchRate}{0.07\%}
\newcommand{\PartialOutcomeCount}{679}
\newcommand{\PartialOutcomeRate}{49.9\%}

\newcommand{\ExpectedEditCount}{202}
\newcommand{\MeanExpectedEdits}{5.05}
\newcommand{\MeanProtectedObjects}{60.55}
\newcommand{\TopModel}{Claude Sonnet 5}
\newcommand{\TopReward}{15.0\%}

\newcommand{\TopRepairProgress}{43.7\%}
\newcommand{\TopUCR}{0.8\%}
\newcommand{\TopValidity}{62.5\%}
\newcommand{\FrontierTopModel}{Claude Sonnet 5}
\newcommand{\FrontierTopReward}{15.0\%}

\newcommand{\FrontierTopRepairProgress}{43.7\%}
\newcommand{\FrontierTopValidity}{62.5\%}
\newcommand{\FrontierTopTruncation}{40.0\%}
\newcommand{\TotalCost}{\$14.88}
\newcommand{\OverallErrorRate}{3.0\%}
\newcommand{\OverallTruncationRate}{37.8\%}
\newcommand{\OverallValidity}{54.0\%}

\title{Vector-Bench: Can Models Surgically Edit SVG Code?}

\author{%
  \begin{tabular}[t]{c@{\qquad\qquad}c}
    Yug Aditi Gupta$^{*}$ & Prannay Hebbar$^{*}$ \\
    \texttt{yug@thetalab.tech} & \texttt{prannay@warping.co} \\
    \textbf{Theta Labs} & \textbf{Warping}
  \end{tabular}\\[6pt]
  \small $^{*}$Equal contribution; the paper and accompanying material are shared work.}

\begin{document}
\maketitle
\raggedbottom

\begin{abstract}
Instruction-based vector editing requires two capabilities: making a requested
change and leaving everything else alone. The second is easy to miss when an
output is judged only as a raster image. We introduce \textbf{Vector-Bench}, a
compact, difficult benchmark of \TaskCount{} SVG repair tasks. Each task pairs a
corrupted SVG program with an author-written visual instruction, a hidden target program,
\MeanExpectedEdits{} annotated repairs on average, and an average of
\MeanProtectedObjects{} protected objects. Instructions describe visible defects
without exposing element identifiers, coordinates, color codes, or path data.
We define a deterministic binary specification reward: requested repairs use
attribute-aware perceptual tolerances, while unrequested rendering- or application-relevant
structure must remain semantically unchanged and the result must be a valid SVG.
Canonical target equality and stricter source fidelity are retained as
diagnostics. Validity-gated repair progress, a near-complete tier, and
valid-output Unintended Change Rate (UCR) explain partial outcomes. We
evaluate \ModelCount{} model endpoints (\OpenModelCount{} listed as open-weight,
\ControlModelCount{} inexpensive controls, and \FrontierModelCount{} frontier
closed endpoints) over \RequestCount{} requests.
The strongest endpoint reaches only \TopReward{} full specification success,
despite \TopRepairProgress{} mean repair progress, showing that apparent repair
progress and specification-faithful editing remain substantially different.
All prompts, outputs, scoring code, costs, and per-task reports are released.
\end{abstract}

\section{Introduction}

An instruction such as ``the amber signal has gone dark; fix it and leave the
station alone'' defines both a positive requirement and a large set of negative
requirements. The signal must change, while the train, clock, cables, platform,
and every unrelated program detail must not. A model can produce a plausible
raster while renaming elements, rewriting path data, shifting an unrelated
object, or deleting metadata. These are not cosmetic differences for an SVG:
they alter an executable, structured artifact~\citep{w3csvg2}.

Most generation metrics emphasize final appearance. That is appropriate for
open-ended synthesis, but insufficient for local editing. A repair benchmark
must answer three separate questions: Did the requested objects change? Did
unmentioned objects remain stable? Is the resulting program valid and usable?
Vector-Bench makes all three observable through hidden target SVGs and explicit
protected-object sets.

\begin{figure*}[t]
  \centering
  \includegraphics[width=0.98\textwidth]{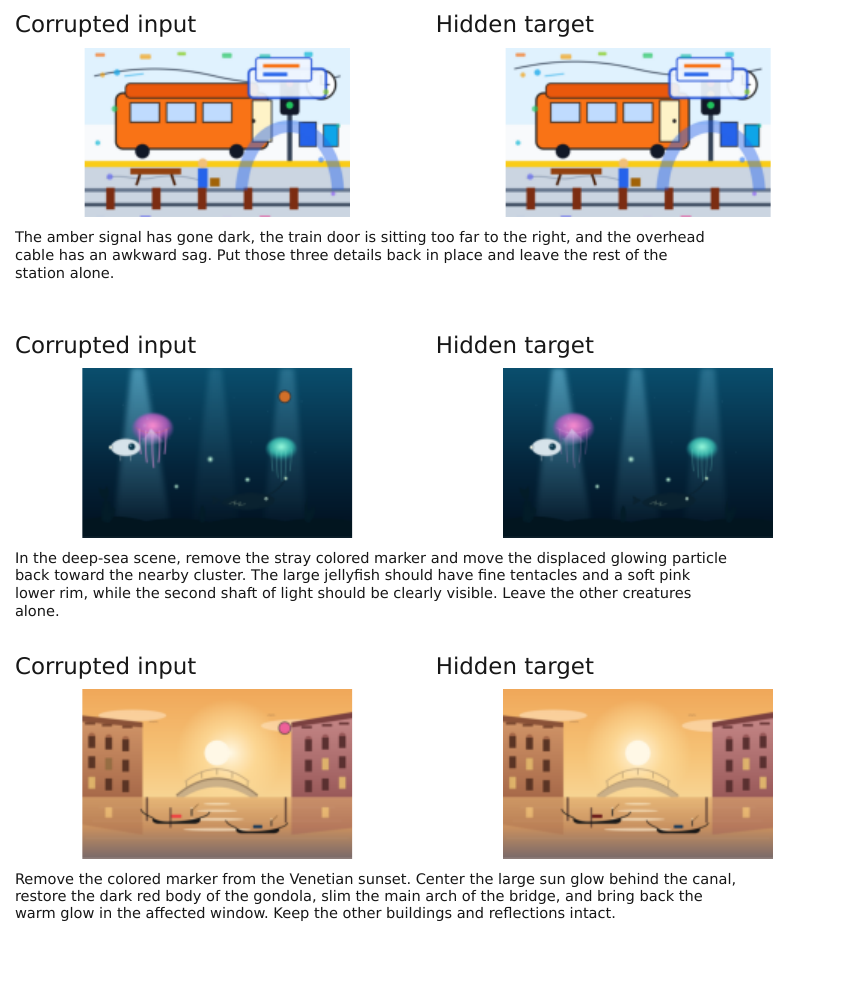}
  \caption{Representative corrupted inputs and hidden targets. Public prompts
  describe visible mistakes in ordinary language without naming source IDs or
  revealing target numeric/path values.}
  \label{fig:examples}
\end{figure*}

The benchmark is deliberately small and dense rather than broad and shallow.
Its \TaskCount{} tasks contain \ExpectedEditCount{} annotated repairs across
station, harbor, campsite, market, and larger scenic illustrations. The public
instruction never names a DOM identifier or serialized numeric/path value. Solvers must
interpret the visible scene and edit the source accordingly. Figure~\ref{fig:examples}
shows representative inputs and targets.

Our contributions are:
\begin{enumerate}
  \item a frozen corpus of dense, author-described SVG repairs with explicit
  intended edits and broad preservation surfaces;
  \item a transparent executable evaluator whose binary specification reward
  combines tolerant requested-edit checks, rendering- and application-relevant semantic
  preservation, and SVG validity, while retaining source fidelity separately;
  \item a reproducible \ModelCount{}-endpoint evaluation with response
  provenance, token usage, cost, failures, and model-produced SVGs; and
  \item an empirical account of the gap between completing visible repairs and
  preserving the full vector program.
\end{enumerate}

\section{Related Work}

\paragraph{Vector generation.}
IconShop generates vector icons from text~\citep{wu2023iconshop}; StarVector
generates SVG code from images and text~\citep{rodriguez2025starvector}; and
LLM4SVG studies complex vector understanding and generation with language
models~\citep{xing2025llm4svg}. These systems establish SVG as a useful meeting
point between visual generation and program synthesis. Vector-Bench focuses on
local repair rather than open-ended generation.

\paragraph{Instruction-based SVG editing.}
SVGEditBench introduced quantitative evaluation for LLM-based SVG editing
\citep{nishina2024svgeditbench}, and SVGEditBench V2 expanded instruction-based
editing evaluation~\citep{nishina2025svgeditbenchv2}. VectorEdits provides a
larger dataset and benchmark for vector-graphics editing
\citep{kuchar2025vectoredits}. Our scope is complementary: we emphasize dense
multi-repair scenes, tolerance-aware repair checks, semantic preservation outside
the requested fields, and publication of every model output. The hidden target
anchors deterministic checks without requiring a model to reproduce every
requested coordinate or color exactly.

\section{Benchmark}
\label{sec:benchmark}

\subsection{Task Format}

Each task is a tuple
\[
  \tau = (x, S_0, S^*, E, U),
\]
where $x$ is an author-written repair instruction, $S_0$ is the corrupted SVG, $S^*$ is
the hidden target, $E$ is a set of expected object-attribute repairs, and $U$ is
the scene-object subset used for object-level preservation diagnostics. The
whole-document preservation gate covers every unrequested node, including nodes
outside $U$. The model receives only $(x,S_0)$.
Target source, expected diffs, target-part lists, and preserve lists are excluded
from the prompt and recorded as hidden evaluator inputs.

Instructions were rewritten after task freezing by the authors. They describe visible objects
and relations (for example, a moon that drifted down and left or a jellyfish
whose tentacles became too heavy) without SVG IDs, numeric coordinates, hex
colors, path commands, or attribute names. A corpus regression hash covers every
non-instruction task field, ensuring this language-only rewrite did not change
inputs, targets, annotations, or ordering.

\subsection{Composition and Difficulty}

The corpus contains 20 compact but busy scenes derived from four authored scene
templates and 20 larger scenic illustrations. Ten tasks are marked hard and 30
very hard. Tasks require between three and nine repairs, including insertion or
deletion, recoloring, displacement, path restoration, line-weight correction,
and opacity restoration. Every unrequested node is covered by the semantic
preservation gate; $U$ enumerates the identified scene objects used for UCR.

\begin{table}[t]
  \centering
  \scriptsize
  \begin{tabular}{lr}
\toprule
Statistic & Value \\
\midrule
Tasks & 40 \\
Annotated repairs & 202 \\
Repairs per task (mean) & 5.05 \\
Protected objects per task (mean) & 60.55 \\
SVG elements per task (mean) & 85.4 \\
Input characters (total) & 294,798 \\
\bottomrule
\end{tabular}

  \caption{Frozen corpus statistics. Element counts include non-rendering SVG
  nodes; protected-object counts refer to identified scene objects.}
  \label{tab:corpus}
\end{table}

\begin{figure}[t]
  \centering
  \includegraphics[width=\columnwidth]{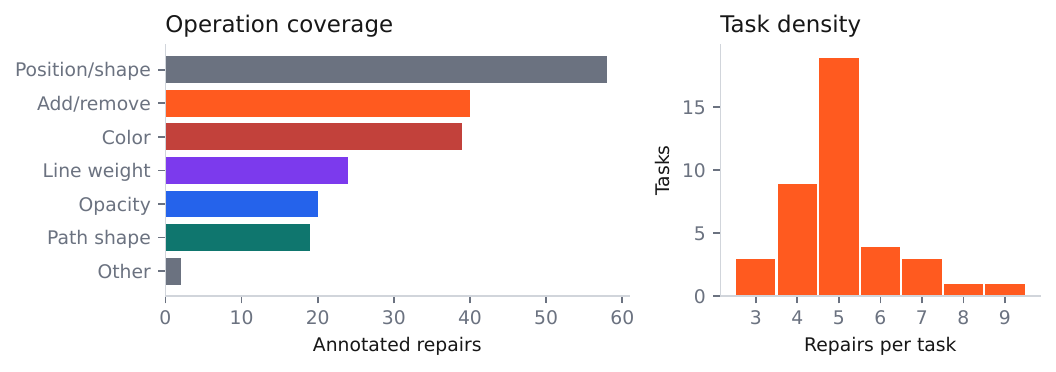}
  \caption{Repair operations and per-task edit density.}
  \label{fig:operations}
\end{figure}

\begin{figure}[t]
  \centering
  \includegraphics[width=\columnwidth]{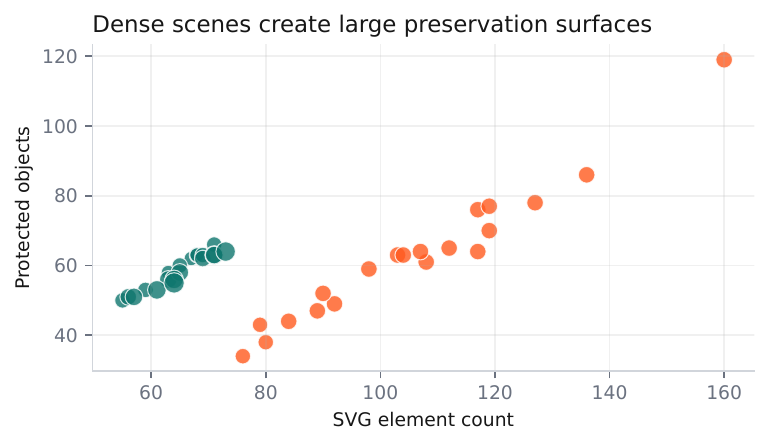}
  \caption{SVG size and preservation surface. Marker area scales with the
  number of requested repairs.}
  \label{fig:complexity}
\end{figure}

\section{Executable Evaluation}
\label{sec:evaluation}

\subsection{Three-Gate Specification Reward}

Let $C(S)$ parse an SVG and canonicalize representation-only variation. We
discard namespace spelling differences and attribute order, normalize numeric
spellings and separators in paths, points, transforms, and view boxes, and map
common equivalent color spellings (for example, \texttt{white},
\texttt{\#fff}, and \texttt{rgb(255,255,255)}) to one form. A second form
$V(S)$ represents rendering- and application-relevant semantics: consistently renamed IDs and
their local references are aligned by tree position, and equivalent inline-style
and presentation-attribute storage is merged. Element order, nesting, geometry,
paint, text, and unresolved references remain strict. Malformed XML has neither
form. Requested objects are resolved by ID first and by stable tree position
when an otherwise equivalent editor has renamed IDs.

For each requested edit $e\in E$, an attribute-aware predicate
$\operatorname{ok}_e(S)$ compares the produced value with the hidden target.
Existence and non-visual values are exact. Scalar geometry uses absolute error;
colors use CIE Lab $\Delta E_{76}$; and point lists use coordinate RMS. Paths
are sampled at 129 length-normalized positions and compared with a symmetric
nearest-point distance combining RMS and the 95th percentile. This admits the
same visible curve expressed with different SVG commands. For matching command
streams, coordinate RMS remains an additional acceptance route. When the
corrupted-to-target distance is $\delta_e$, the tolerance is bounded by
\[
  t_e=\min(c_e,\max(f_e,r_e\delta_e),0.9\delta_e),
\]
so the known corrupted value cannot pass. Scalar checks use $r_e=0.50$, color
uses 0.45, point lists use 0.55, and paths use 0.70. Position caps are 2.5\% of
the relevant viewport dimension, color is capped at 18 $\Delta E_{76}$, opacity
at 0.12, and path/point geometry at 3\% of the smaller viewport dimension.
Stroke width is capped at one user unit or 40\% of target width, whichever is
larger. Floors avoid rejecting immaterial numeric noise. Every per-check report
exposes target distance, tolerance, baseline corruption distance, and progress.

Approximation is permitted only at requested fields. Let $M_E$ mask those fields
in both the produced and target trees (and remove requested insertion/deletion
sites). The primary preservation gate requires
$V(M_E(S))=V(M_E(S^*))$. It accepts consistent ID renaming and style-storage
rewrites but catches changes to unlisted definitions, anonymous nodes, root
attributes, element order, and unrequested visual attributes. We separately
report the stricter source predicate $C(M_E(S))=C(M_E(S^*))$. The validity gate
requires a complete SVG root, well-formed XML, unique nonempty IDs, valid path,
point-list, transform, and view-box syntax, and resolvable local URL and
\texttt{href} references.

The primary reward remains binary:
\[
  \begin{aligned}
  R_{\mathrm{spec}}(S)={}&\mathbb{1}[\operatorname{valid}(S)]\\
  &\cdot\mathbb{1}[\forall e\in E:\operatorname{ok}_e(S)]\\
  &\cdot\mathbb{1}[V(M_E(S))=V(M_E(S^*))].
  \end{aligned}
\]
Canonical target equality $\mathbb{1}[C(S)=C(S^*)]$ and source preservation are
diagnostics and are not required for a pass. No learned judge or model-based
grader appears in the public protocol.

\begin{table*}[t]
  \centering
  \small
  \begin{tabular}{lrrrrrr}
\toprule
Control & Full & Repair & Progress & Clean & UCR & Valid \\
\midrule
Return corrupted input & 0.0 & 0.0 & 0.0 & 100.0 & 0.0 & 100.0 \\
Hidden target copy & 100.0 & 100.0 & 100.0 & 100.0 & 0.0 & 100.0 \\
Target + protected mutation & 0.0 & 100.0 & 100.0 & 0.0 & 1.7 & 100.0 \\
Malformed target & 0.0 & 0.0 & 0.0 & 0.0 & -- & 0.0 \\
\bottomrule
\end{tabular}

  \caption{Deterministic evaluator controls over all \TaskCount{} tasks. Values
  are percentages; UCR is conditional on valid SVGs and is undefined when none
  are valid. The protected-mutation control verifies that completing every
  repair cannot compensate for an unrequested change.}
  \label{tab:controls}
\end{table*}

\subsection{Evaluator Audit}
\label{sec:evaluator-audit}

The evaluator is tested as an executable artifact rather than treated as an
unexamined source of labels. The target SVG must receive reward one; the
corrupted input must fail at least one requested repair; and unintended
deletions, unrequested attribute changes, duplicate identifiers, broken
references, malformed paths, and unknown added elements must fail the
preservation or validity gate. Regression tests also cover approximate numeric
and color repairs, alternate path command topologies, equivalent style storage,
and consistent identifier renaming. The frozen corpus has a regression hash
over every non-instruction task field. These checks establish implementation
invariants, not human perceptual validity: the comparator's scope and the
absence of a human agreement study remain limitations.

Table~\ref{tab:controls} reports four deterministic controls over all tasks.
Returning the corrupted input remains valid and perfectly preserved but earns
no repair credit. Copying the hidden target is the evaluator upper bound. Adding
one protected metadata mutation to each target preserves all requested repairs
but forces the binary reward to zero, and malformed targets receive neither
validity nor progress credit. These are evaluator controls, not solver baselines:
the target is unavailable to evaluated models.

\subsection{Diagnostic Metrics}

For expected checks $E$, tolerance-aware edit completion is
\[
  \mathrm{Edit}(S)=\frac{1}{|E|}\sum_{e\in E}\mathbb{1}[\operatorname{ok}_e(S)].
\]
For valid outputs and checks with a distance, repair progress is
$q_e=\operatorname{clip}(1-d_e/\delta_e,0,1)$; exact and existence checks use
binary progress. Invalid outputs set every $q_e$ to zero, preventing a missing
document from receiving deletion credit. We report $|E|^{-1}\sum_e q_e$ in addition to completion. A
\emph{near-complete} output is valid and semantically preserved, misses at most
one requested check, and has at least 80\% mean progress. It still receives zero
binary reward.

For protected objects $U$, preservation and unintended-change rate are
\[
  \begin{aligned}
  \mathrm{Pres}(S)&=\frac{1}{|U|}\sum_{u\in U}\mathbb{1}[V(u_S)=V(u_{S^*})],\\
  \mathrm{UCR}(S)&=1-\mathrm{Pres}(S).
  \end{aligned}
\]
We additionally report all-repairs pass rate, whole-document semantic-clean
rate, source-preservation rate, SVG validity, whitespace-normalized source-string
equality to the target, canonical target match, provider errors, scheduled
elapsed time, tokens, and cost.
Model-level UCR is averaged over valid outputs so truncation is not mislabeled
as collateral editing; it is undefined when a model has no valid output. A
provider error or invalid output receives reward and repair progress zero.

\begin{table*}[t]
  \centering
  \small
  \begin{tabular}{lrrrrrrrr}
\toprule
Model & Full $\uparrow$ & Near $\uparrow$ & Repair $\uparrow$ & Progress $\uparrow$ & Clean $\uparrow$ & UCR $\downarrow$ & Valid $\uparrow$ & Cost \\
\midrule
Claude Sonnet 5 & 15.0 & 2.5 & 15.0 & 43.7 & 42.5 & 0.8 & 62.5 & \$2.032 \\
KAT Coder Air V2.5 & 12.5 & 0.0 & 15.0 & 27.3 & 25.0 & 0.3 & 42.5 & \$0.128 \\
MiniMax M3 & 10.0 & 0.0 & 10.0 & 33.0 & 25.0 & 1.0 & 45.0 & \$0.240 \\
HY 3 & 7.5 & 2.5 & 10.0 & 56.8 & 37.5 & 1.8 & 100.0 & \$0.092 \\
Gemma 4 31B IT & 5.0 & 5.0 & 10.0 & 54.8 & 47.5 & 0.8 & 87.5 & \$0.084 \\
Qwen3.5 397B A17B & 5.0 & 0.0 & 5.0 & 39.8 & 32.5 & 1.4 & 65.0 & \$0.618 \\
DeepSeek V4 Pro & 5.0 & 2.5 & 5.0 & 30.6 & 20.0 & 1.2 & 47.5 & \$0.508 \\
Kimi K2.6 & 5.0 & 5.0 & 5.0 & 16.0 & 15.0 & 0.3 & 25.0 & \$1.077 \\
Qwen3 Coder & 2.5 & 0.0 & 5.0 & 54.7 & 45.0 & 1.4 & 100.0 & \$0.295 \\
Gemini 3.1 Flash Lite & 2.5 & 0.0 & 7.5 & 54.5 & 25.0 & 2.5 & 100.0 & \$0.269 \\
\bottomrule
\end{tabular}

  \caption{Top ten endpoints under binary specification reward. Full, Near,
  Repair, Progress, Clean, UCR, and Valid are percentages; UCR is conditional
  on valid outputs, and cost is provider-reported for all 40 tasks.
  Full results are in Appendix~\ref{sec:full-results}.}
  \label{tab:main}
\end{table*}

\section{Model Study}
\label{sec:study}

\subsection{Protocol}

We evaluate \OpenModelCount{} endpoints designated as open-weight in the study
manifest, \ControlModelCount{} inexpensive closed controls, and
\FrontierModelCount{} frontier closed endpoints through the OpenRouter
chat-completions API~\citep{openrouter2026}. These designations record the
endpoint grouping used for analysis and are not an independent license audit.
The exact endpoint IDs appear in Appendix~\ref{sec:full-results}.

Every model receives the same system instruction, author-written repair request, and
corrupted SVG source. We record one scored outcome per task and omit temperature
rather than imposing a provider-incompatible value. In this run, input-adaptive
output ceilings range from 4,096 to 7,002 tokens. The runner retries rate
limits and transient server failures against the same endpoint. It never falls
back to another model; the client permits two transport retries inside each
harness attempt. We record requested and resolved model IDs, response ID,
finish reason, prompt, completion and reasoning tokens when supplied, scheduled
wall-clock elapsed time, and provider-reported cost. Elapsed time begins when a
model--task job is scheduled and therefore includes local concurrency-queue wait;
we expose it for run auditing, not as endpoint latency. Each source cohort uses
a reservation-based budget guard capped at \$25; the completed combined study
cost \TotalCost{}. The current runner additionally emits credential-redacted,
append-only request, retry, response, extraction, and scoring events. Because the
reported study predates this complete attempt logger, its public rows are marked
\texttt{legacy\_final\_record}: the extracted final SVG (or malformed final body)
and metadata are retained, but discarded retry envelopes and response wrappers
cannot be reconstructed.

This is an endpoint comparison, not a general ranking of model intelligence.
It uses one decoded response per model--task pair and does not include a human
performance baseline. The released artifact supports exact reruns of the
scored protocol, but does not estimate sampling variance, human performance, or
performance on unseen SVG distributions.

\subsection{Uncertainty}

We report two-sided 95\% Wilson score intervals over the 40 binary task outcomes
for each model. Unlike a nonparametric bootstrap, Wilson intervals remain
informative when a model records zero passes. They quantify uncertainty under a
Bernoulli task-sampling interpretation only; they do not estimate decoding
variance because each model--task pair has one scored outcome. Endpoint tables
are ordered by full specification pass, mean repair progress, lower valid-output
UCR, validity, and model name.

\section{Results}
\label{sec:results}

\begin{figure*}[t]
  \centering
  \includegraphics[width=0.84\textwidth]{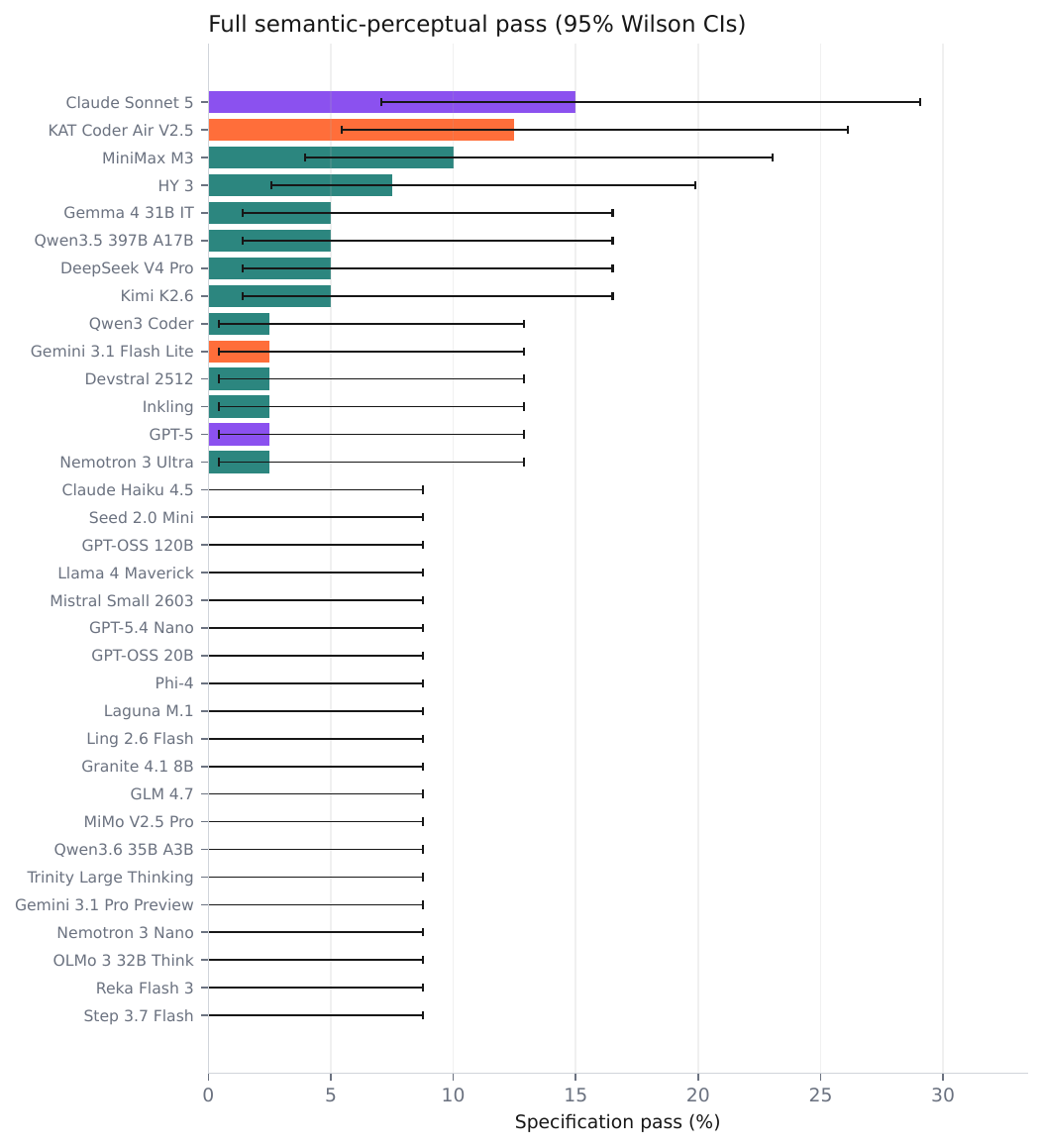}
  \caption{Binary specification success for all endpoints. Orange bars are inexpensive
  controls, teal bars are endpoints listed as open-weight, and purple bars are
  frontier closed endpoints.}
  \label{fig:reward}
\end{figure*}

\paragraph{Specification success remains rare.}
Only \SpecificationPassCount{} of \RequestCount{} outcomes satisfy all three
gates (\SpecificationPassRate{}), while \NearPassCount{} (\NearPassRate{}) are
near-complete and \PartialOutcomeCount{} (\PartialOutcomeRate{}) make at least
one annotated repair. Validity-gated repair progress is \OverallRepairProgress{}
over all scheduled outcomes and \ValidRepairProgress{} conditional on valid
outputs. Valid-output UCR is \OverallValidUCR{} overall.
Across all outcomes, \RepairPassRate{} satisfy every requested repair and
\CleanPassRate{} preserve the masked document semantically; the stricter source
gate passes \SourceCleanPassRate{}, and only \TargetMatchRate{} canonically
match the full hidden target. The strongest endpoint, \TopModel{}, obtains
\TopReward{} specification success. Its repair progress is substantially higher
at \TopRepairProgress{}, while its valid-output UCR is \TopUCR{} and SVG validity is
\TopValidity{}. The difference is the central benchmark result: models frequently
perform some requested visual work but fail the full repair-and-preserve
contract. Figure~\ref{fig:reward} reports Wilson intervals for every
endpoint.

\paragraph{Frontier endpoints remain subject to the same bottlenecks.}
Within the frontier cohort, \FrontierTopModel{} ranks highest with
\FrontierTopReward{} specification success and \FrontierTopRepairProgress{} repair
progress. Its SVG validity is \FrontierTopValidity{} and its truncation rate
is \FrontierTopTruncation{}. Length-limited outcomes elsewhere in the cohort
remain failures because the protocol requires a complete usable artifact.

\begin{figure*}[t]
  \centering
  \includegraphics[width=0.82\textwidth]{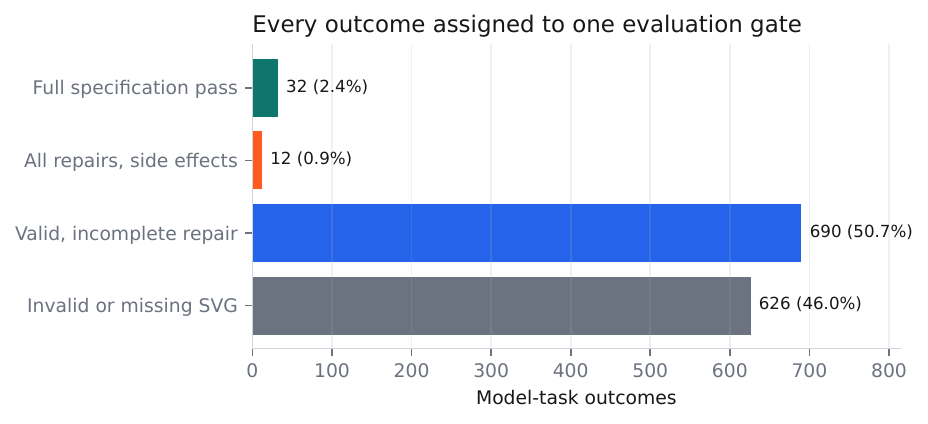}
  \caption{Mutually exclusive decomposition of all \RequestCount{} outcomes.
  The first two rows distinguish clean specification success from completed
  repairs with side effects; the remaining rows isolate incomplete valid edits
  from invalid or missing artifacts.}
  \label{fig:gates}
\end{figure*}

\paragraph{The failure mass is mostly incomplete or unusable output.}
Figure~\ref{fig:gates} partitions every model--task pair into exactly one gate
outcome. This decomposition prevents low binary success from being interpreted
as one homogeneous failure mode: a valid partial repair, an all-repairs output
with collateral changes, and an invalid or missing SVG all receive zero reward
for different reasons.

\paragraph{The conclusion is stable to moderate tolerance changes.}
Table~\ref{tab:sensitivity} rescales every distance-based tolerance while
retaining exact structural checks, validity, semantic preservation, and the
rule that the known corruption cannot pass. The absolute pass rate changes, as
expected, but remains low under both stricter and looser settings. Thus the
result is not created by requiring canonical target equality, although the
precise binary rate remains calibration-dependent.

\begin{table}[t]
  \centering
  \small
  \begin{tabular}{lrrrr}
\toprule
Scale & Full $n$ & Full \% & Repair $n$ & Repair \% \\
\midrule
$0.5\times$ & 11 & 0.81 & 13 & 0.96 \\
$\mathbf{1.0\times}$ & 32 & 2.35 & 44 & 3.24 \\
$1.5\times$ & 46 & 3.38 & 67 & 4.93 \\
\bottomrule
\end{tabular}

  \caption{Tolerance sensitivity over the same \RequestCount{} recorded
  outcomes; the primary setting is bold. Rates are percentages. Scaling applies only to distance-based
  requested-edit tolerances; all other reward gates are unchanged.}
  \label{tab:sensitivity}
\end{table}

\begin{figure*}[t]
  \centering
  \begin{minipage}[t]{0.48\textwidth}
    \centering
    \includegraphics[width=\linewidth]{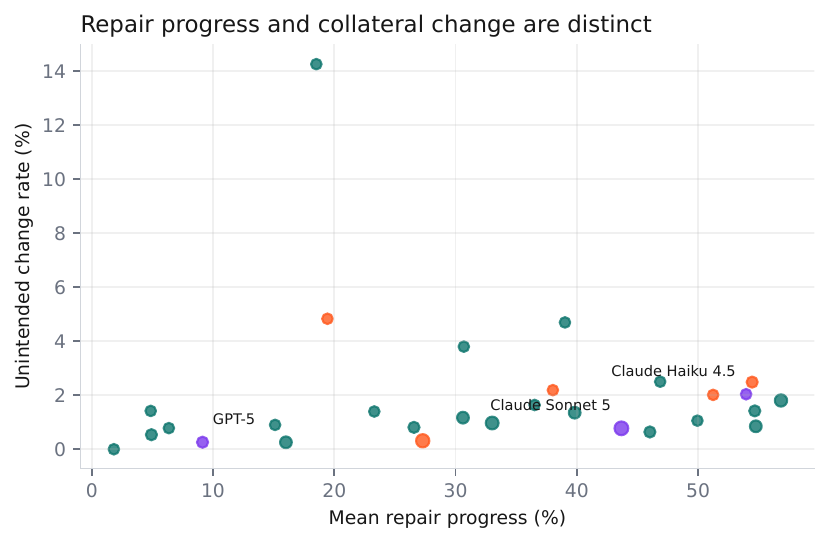}
    \small (a) Repair progress versus unintended change.
  \end{minipage}\hfill
  \begin{minipage}[t]{0.48\textwidth}
    \centering
    \includegraphics[width=\linewidth]{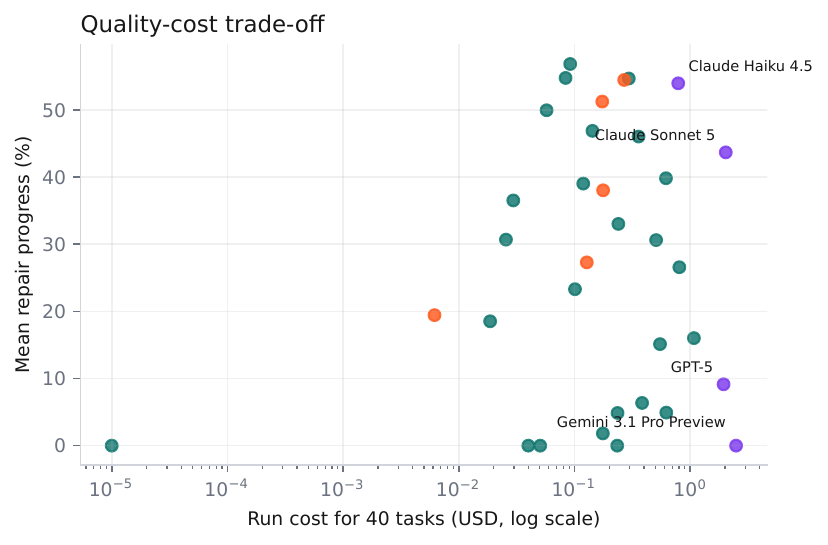}
    \small (b) Repair progress versus provider-reported cost.
  \end{minipage}
  \caption{Complementary model-level diagnostics. In (a), marker area scales
  with specification pass rate and lower right is preferable. Cost in (b) is
  the recorded total for 40 tasks on a logarithmic axis.}
  \label{fig:tradeoffs}
\end{figure*}

\paragraph{Repair and preservation are separate axes.}
Figure~\ref{fig:tradeoffs} (left) plots mean repair progress against valid-output UCR. A model can
move right by correcting visible defects while also moving upward by rewriting
protected scene objects. Reporting completion alone would hide this behavior;
reporting only full success would hide useful partial capability.

\begin{figure*}[t]
  \centering
  \includegraphics[width=0.88\textwidth]{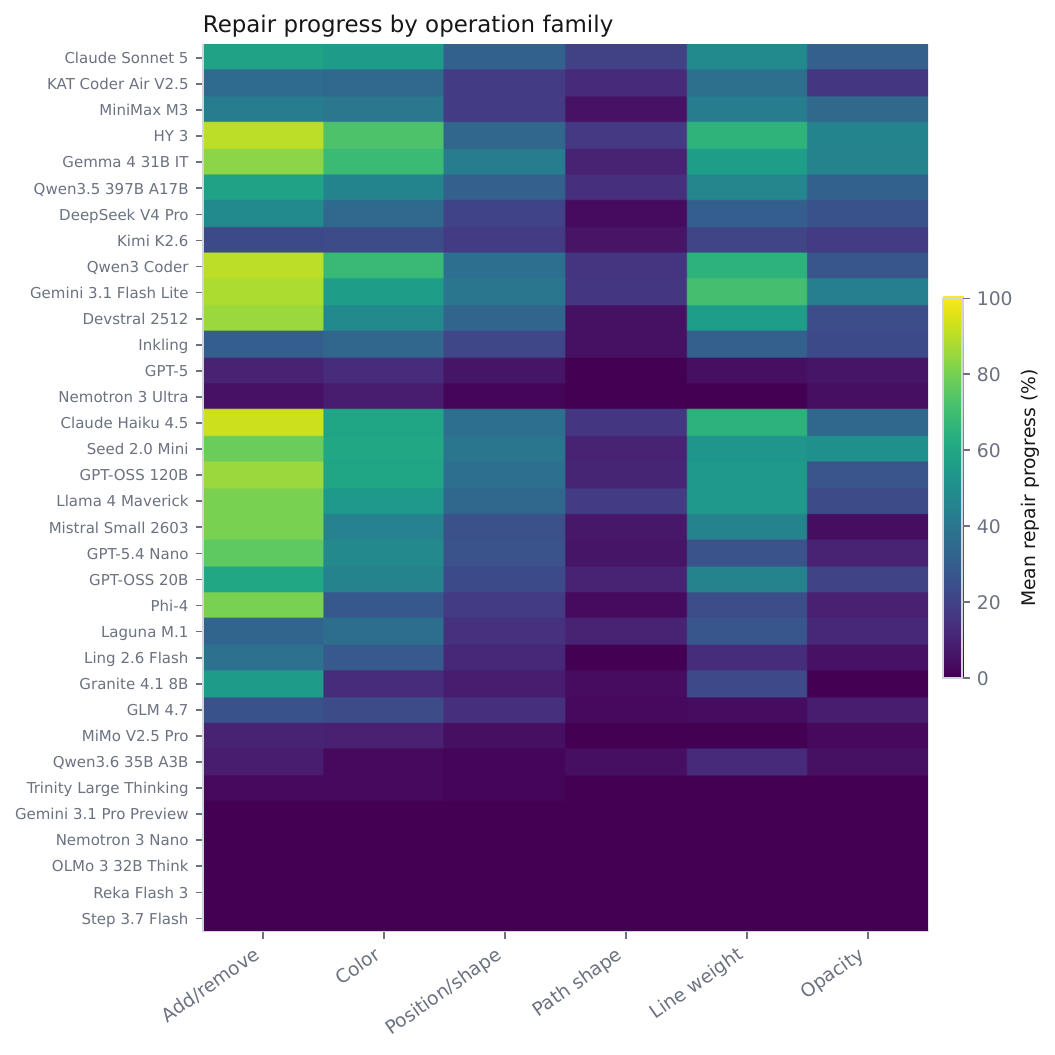}
  \caption{Mean expected-edit repair progress by operation family. Rows follow overall
  specification-pass rank.}
  \label{fig:heatmap}
\end{figure*}

\paragraph{Operation families expose different bottlenecks.}
Figure~\ref{fig:heatmap} decomposes the expected checks. Simple color or
add/remove corrections need not predict path-shape or position restoration,
especially because the prompt gives visual descriptions instead of literal DOM
values. This decomposition also distinguishes an endpoint that attempted an edit
but selected the wrong source element from one that returned an invalid or
unchanged document.

\paragraph{Cost is not a sufficient capability proxy.}
The run spans inexpensive small endpoints and larger reasoning or coding models.
Figure~\ref{fig:tradeoffs} (right) shows repair progress against total 40-task cost. It is a
descriptive endpoint comparison rather than a hardware-efficiency claim:
OpenRouter pricing and routing can change, and the study uses one scored outcome
per task.

\section{Error Analysis}
\label{sec:error-analysis}

Appendix~\ref{sec:trace-example} traces a full pass and
Appendix~\ref{sec:trace} a full failure; both make the failure modes below
concrete.

\paragraph{Literal partial repair.}
Models often identify an obvious semantic defect, such as a signal with the wrong
color, while leaving a displaced object or malformed curve unresolved. These
outputs earn partial edit-completion credit but specification reward zero.

\paragraph{Global rewrites.}
Some responses regenerate or reformat substantial portions of the SVG. The
semantic normalizer accepts harmless formatting, equivalent style storage, and
consistent ID renaming, but changed visible paths, reordered scene objects,
broken references, or missing definitions count as changes. The strict
source-fidelity diagnostic still surfaces source rewrites when their visual
semantics are preserved.

\paragraph{Approximate repair without collateral change.}
Prompts intentionally omit exact coordinates, color codes, and path commands.
The tolerance-aware repair gate accepts bounded alternatives to the hidden target,
but only when the remainder of the SVG is unchanged. This separates a reasonable
visual approximation from a global rewrite. The full-target match diagnostic
quantifies canonical reproduction without controlling the primary reward.

\paragraph{Provider and truncation failures.}
Only \OverallValidity{} of scored outcomes pass structural SVG validity. The harness
preserves raw malformed output, API errors, and finish reasons in the released
records. Transient failures are retried only against the requested endpoint;
persistent failures remain zero-scored observations. The aggregate provider
error rate is \OverallErrorRate{}, and \OverallTruncationRate{} of all outcomes
end at the provider's output-length limit; parse failure is measured separately.

\section{Limitations and Artifact Risks}

Vector-Bench has only \TaskCount{} tasks and should be treated as a focused
stress test, not a comprehensive measure of visual editing. Public targets make
future contamination possible. The deterministic tolerances are transparent but
cannot represent every perceptually acceptable alternative. Path comparison uses
finite geometric samples and can miss narrow or highly localized differences;
CIE76 is only an approximate color metric, and viewport-relative caps may be too
strict or permissive for particular scenes. The semantic comparator does not
execute arbitrary CSS, browser layout, animation, scripting, accessibility, or
editor-specific behavior.

Instructions and acceptable edits were authored and checked by the benchmark
authors; no independent annotator-agreement or human-performance study was run.
The sensitivity analysis varies tolerance width but is not a substitute for
human calibration. Consequently, binary reward should be interpreted as success
under this released executable specification, not universal perceptual approval.

The evaluation uses one scored outcome per model--task pair and therefore cannot
characterize sampling stability. OpenRouter can change prices, routing, model versions, and
availability; requested and resolved IDs are published to make this visible.
The ``open-weight'' grouping is manifest metadata, not a legal determination.

Twenty scenic source files in the frozen corpus lack recoverable per-file
provenance metadata. We do not claim original authorship of those illustrations.
They are retained to preserve the evaluated task set, marked as provenance
pending in the artifact, and should be treated as research-only until provenance
is resolved. This limits redistribution confidence and is a material artifact
risk.

Finally, the benchmark evaluates text-and-source editing rather than multimodal
screen interaction. Models see SVG code and a visual description, but not a
raster preview or interactive editor. Performance therefore combines scene
reasoning, source localization, and code generation under one protocol.

\section{Conclusion}

Vector-Bench isolates a practical requirement for editing agents: requested
change is not enough; preservation is part of the specification. Author-written
visual instructions make the task less like copying a patch, while executable SVG
targets make evaluator decisions inspectable. Across \ModelCount{} endpoints, partial
repair is much more common than specification-faithful repair. The released
outputs and diff reports make that gap inspectable at the object and source level
and provide a compact target for improving structured visual editors.

\bibliography{custom}

\appendix
\clearpage
\onecolumn
\section{Artifact Notes}

\paragraph{Reproducibility.}
The release includes the frozen corpus, prompts, runner, evaluator, tests,
analysis, paper source, Harbor tasks, and a chronological per-output trace viewer.
Credentials and authorization headers are excluded. The released study includes
sanitized final-response records and verifier snapshots; unavailable legacy retry
and transport envelopes are explicitly labeled rather than inferred. New runs
retain these events append-only. The public artifact is available at
\url{https://www.vecbench.xyz}, with source at
\url{https://github.com/yug-space/vector-edit-gym}.

\paragraph{Shared authorship.}
Yug Aditi Gupta and Prannay Hebbar contributed equally to the benchmark,
artifact, analysis, and writing. The paper and accompanying material are shared
work.

\section{Full Endpoint Results}
\label{sec:full-results}

Values are percentages. Errors include persistent provider failures and missing
usable SVG output. Endpoint IDs, resolved IDs, response IDs, and per-task details
are available in the machine-readable release.

\scriptsize
\begin{longtable}{llrrrrrrrrr}
\toprule
Model & Group & Full & Near & Repair & Progress & Clean & UCR & Valid & Trunc. & Errors \\
\midrule
\endhead
Claude Sonnet 5 & frontier & 15.0 & 2.5 & 15.0 & 43.7 & 42.5 & 0.8 & 62.5 & 40.0 & 0.0 \\
KAT Coder Air V2.5 & cheap-control & 12.5 & 0.0 & 15.0 & 27.3 & 25.0 & 0.3 & 42.5 & 57.5 & 0.0 \\
MiniMax M3 & open-weight & 10.0 & 0.0 & 10.0 & 33.0 & 25.0 & 1.0 & 45.0 & 55.0 & 0.0 \\
HY 3 & open-weight & 7.5 & 2.5 & 10.0 & 56.8 & 37.5 & 1.8 & 100.0 & 0.0 & 0.0 \\
Gemma 4 31B IT & open-weight & 5.0 & 5.0 & 10.0 & 54.8 & 47.5 & 0.8 & 87.5 & 5.0 & 0.0 \\
Qwen3.5 397B A17B & open-weight & 5.0 & 0.0 & 5.0 & 39.8 & 32.5 & 1.4 & 65.0 & 32.5 & 0.0 \\
DeepSeek V4 Pro & open-weight & 5.0 & 2.5 & 5.0 & 30.6 & 20.0 & 1.2 & 47.5 & 52.5 & 0.0 \\
Kimi K2.6 & open-weight & 5.0 & 5.0 & 5.0 & 16.0 & 15.0 & 0.3 & 25.0 & 72.5 & 2.5 \\
Qwen3 Coder & open-weight & 2.5 & 0.0 & 5.0 & 54.7 & 45.0 & 1.4 & 100.0 & 0.0 & 0.0 \\
Gemini 3.1 Flash Lite & cheap-control & 2.5 & 0.0 & 7.5 & 54.5 & 25.0 & 2.5 & 100.0 & 0.0 & 0.0 \\
Devstral 2512 & open-weight & 2.5 & 5.0 & 2.5 & 46.0 & 57.5 & 0.6 & 97.5 & 0.0 & 0.0 \\
Inkling & open-weight & 2.5 & 0.0 & 2.5 & 26.6 & 27.5 & 0.8 & 45.0 & 55.0 & 0.0 \\
GPT-5 & frontier & 2.5 & 0.0 & 2.5 & 9.1 & 7.5 & 0.3 & 15.0 & 85.0 & 0.0 \\
Nemotron 3 Ultra & open-weight & 2.5 & 0.0 & 2.5 & 4.9 & 5.0 & 0.5 & 7.5 & 92.5 & 0.0 \\
Claude Haiku 4.5 & frontier & 0.0 & 0.0 & 2.5 & 54.0 & 27.5 & 2.0 & 100.0 & 0.0 & 0.0 \\
Seed 2.0 Mini & cheap-control & 0.0 & 2.5 & 2.5 & 51.2 & 35.0 & 2.0 & 92.5 & 2.5 & 0.0 \\
GPT-OSS 120B & open-weight & 0.0 & 5.0 & 0.0 & 49.9 & 42.5 & 1.1 & 95.0 & 5.0 & 0.0 \\
Llama 4 Maverick & open-weight & 0.0 & 0.0 & 2.5 & 46.9 & 20.0 & 2.5 & 92.5 & 0.0 & 0.0 \\
Mistral Small 2603 & open-weight & 0.0 & 0.0 & 0.0 & 39.0 & 5.0 & 4.7 & 95.0 & 0.0 & 0.0 \\
GPT-5.4 Nano & cheap-control & 0.0 & 0.0 & 0.0 & 38.0 & 22.5 & 2.2 & 97.5 & 0.0 & 0.0 \\
GPT-OSS 20B & open-weight & 0.0 & 2.5 & 2.5 & 36.5 & 27.5 & 1.6 & 72.5 & 27.5 & 0.0 \\
Phi-4 & open-weight & 0.0 & 0.0 & 0.0 & 30.7 & 12.5 & 3.8 & 92.5 & 0.0 & 0.0 \\
Laguna M.1 & open-weight & 0.0 & 2.5 & 0.0 & 23.3 & 25.0 & 1.4 & 42.5 & 55.0 & 0.0 \\
Ling 2.6 Flash & cheap-control & 0.0 & 0.0 & 0.0 & 19.4 & 15.0 & 4.8 & 87.5 & 5.0 & 0.0 \\
Granite 4.1 8B & open-weight & 0.0 & 0.0 & 0.0 & 18.5 & 10.0 & 14.3 & 77.5 & 2.5 & 0.0 \\
GLM 4.7 & open-weight & 0.0 & 0.0 & 0.0 & 15.1 & 12.5 & 0.9 & 30.0 & 70.0 & 0.0 \\
MiMo V2.5 Pro & open-weight & 0.0 & 0.0 & 0.0 & 6.4 & 5.0 & 0.8 & 10.0 & 87.5 & 0.0 \\
Qwen3.6 35B A3B & open-weight & 0.0 & 0.0 & 2.5 & 4.9 & 2.5 & 1.4 & 7.5 & 92.5 & 0.0 \\
Trinity Large Thinking & open-weight & 0.0 & 0.0 & 0.0 & 1.8 & 0.0 & 0.0 & 2.5 & 97.5 & 0.0 \\
Gemini 3.1 Pro Preview & frontier & 0.0 & 0.0 & 0.0 & 0.0 & 0.0 & -- & 0.0 & 100.0 & 0.0 \\
Nemotron 3 Nano & open-weight & 0.0 & 0.0 & 0.0 & 0.0 & 0.0 & -- & 0.0 & 80.0 & 0.0 \\
OLMo 3 32B Think & open-weight & 0.0 & 0.0 & 0.0 & 0.0 & 0.0 & -- & 0.0 & 0.0 & 100.0 \\
Reka Flash 3 & open-weight & 0.0 & 0.0 & 0.0 & 0.0 & 0.0 & -- & 0.0 & 12.5 & 0.0 \\
Step 3.7 Flash & open-weight & 0.0 & 0.0 & 0.0 & 0.0 & 0.0 & -- & 0.0 & 100.0 & 0.0 \\
\bottomrule
\end{longtable}

\small
\clearpage

\section{Representative Output Trace}
\label{sec:trace-example}

Table~\ref{tab:trace-example} follows one published outcome through request,
extraction, tolerant repair checks, preservation, validity, and binary reward.
The example is useful because the generated handlebar is not the canonical path,
yet its sampled geometry lies inside the released tolerance. Thus an acceptable
approximation passes without weakening the requirement that unrequested content
remain unchanged.

\begin{table}[h]
\centering
\small
\begin{tabular}{p{0.18\linewidth}p{0.76\linewidth}}
\toprule
Trace event & Recorded value \\
\midrule
Identity & Task \texttt{sv\_005}; requested and resolved endpoint
\texttt{anthropic/claude-sonnet-5}; trace
\texttt{legacy:anthropic/claude-sonnet-5:sv\_005}. \\
Instruction & ``A stray price sticker has appeared in the market, one pear is
red instead of green, and the bicycle handlebar has a distorted bend. Clean up
those three issues while keeping the stalls and shoppers unchanged.'' \\
Request & System repair protocol plus the corrupted SVG; provider-default
temperature; 4,243-token output ceiling. The request used 3,700 prompt tokens. \\
Response & Response \texttt{gen-1784563591-dtcYmLQV0qHEuDb7ucuZ}; stop reason
\texttt{stop}; 3,521 completion tokens; recorded cost \$0.04261. \\
Extraction & A complete SVG was extracted. This historical row is labeled
\texttt{legacy\_final\_record}: the extracted SVG is retained, but the original
response wrapper and retry envelopes were not retained and are not reconstructed. \\
Handlebar check & Expected path
\texttt{M122 96 C130 88 136 92 136 98}; produced path
\texttt{M122 96 C128 90 132 90 134 96}. Sampled-geometry distance
$1.883 \leq 2.961$; check passed; repair progress 55.5\%. \\
Pear check & Expected and produced fill \texttt{\#a3e635};
$\Delta E_{76}=0 \leq 18$; check passed; repair progress 100\%. \\
Sticker check & Expected absence and produced absence; check passed; repair
progress 100\%. \\
Preservation & Every protected object passed semantic comparison;
preservation 100\%; valid-output UCR 0\%. \\
Validity & The output parsed and passed standalone SVG validity checks. \\
Outcome & All three required repairs passed, semantic preservation passed, and
validity passed. Binary reward $=1$; mean soft repair progress $=85.2\%$. \\
\bottomrule
\end{tabular}
\caption{Chronological verifier trace for a published full pass. Soft progress
retains information about approximate repair magnitude; it is not an additional
binary gate once each requested edit is within tolerance.}
\label{tab:trace-example}
\end{table}

The interactive release exposes the corresponding prompt, response artifact,
per-edit distances, preservation report, and score history at
\url{https://www.vecbench.xyz/traces/run} using task \texttt{sv\_005} and model
\texttt{anthropic/claude-sonnet-5}.

\section{A Worked Failure Trace}
\label{sec:trace}

\begin{figure*}[h]
  \centering
  \includegraphics[width=0.92\textwidth]{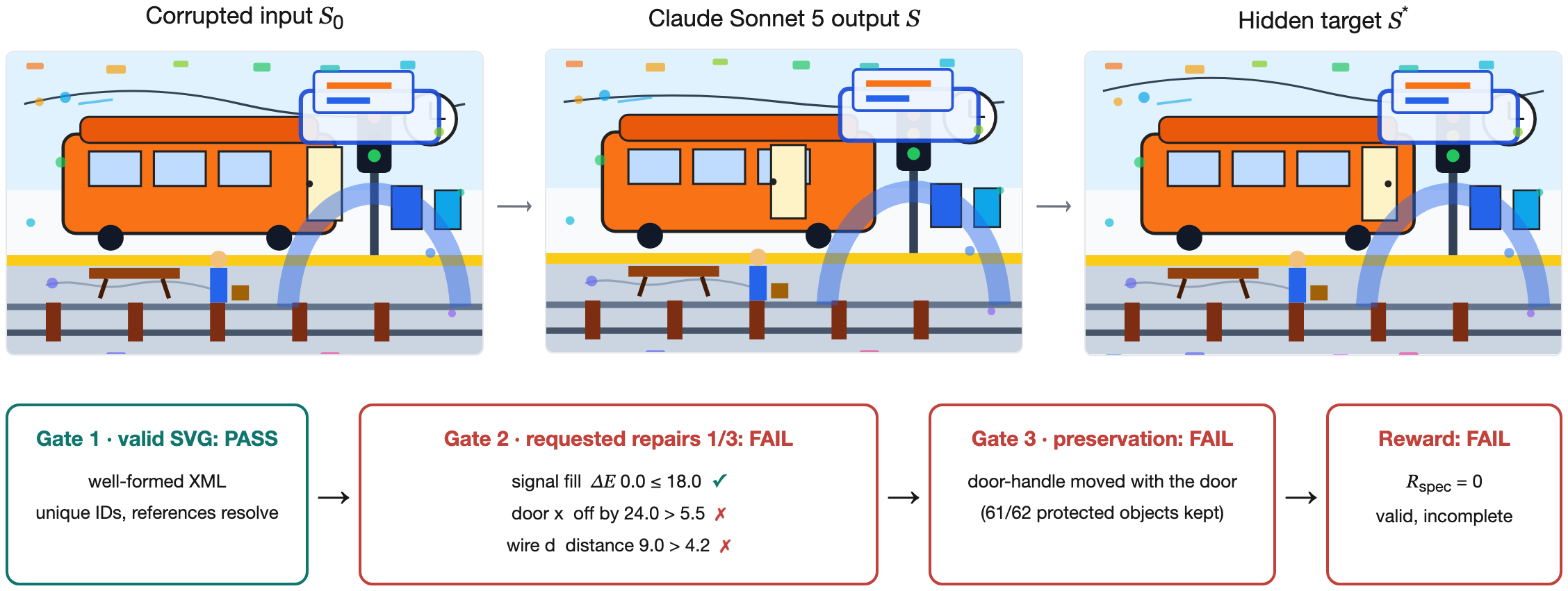}
  \caption{One scored trace, end to end. Top: corrupted input, model output,
  hidden target. Bottom: the three-gate evaluation. The document is valid
  (gate 1). Only the signal-fill repair meets its tolerance (gate 2). Moving
  the door handle changes a protected object (gate 3). The specification
  reward is zero.}
  \label{fig:trace}
\end{figure*}

The trace above shows a pass; this one shows why passes are rare. In
Figure~\ref{fig:trace}, the strongest endpoint in the study, Claude Sonnet 5,
answers the station task this paper opens with (\texttt{sv\_001}). The output
looks plausible and is valid SVG. But only the signal is repaired
($\Delta E_{76}=0$). The door overshoots the hidden target by 24 user units.
The reshaped wire lands farther from the target than the corruption it was
asked to fix. The model also drags the door handle, a protected object the
corruption never touched, along with the door. One repair out of three plus
one collateral change put the response in the ``valid, incomplete repair''
bucket that dominates Figure~\ref{fig:gates}. The reward is zero. This is the
paper's central gap in a single trace: visible repair work, not the
repair-and-preserve contract. The failure modes of
Section~\ref{sec:error-analysis} co-occur in one response. The released
viewer exposes every per-check distance, tolerance, and rendered
before/after.\footnote{\url{https://www.vecbench.xyz/traces/run?task=sv_001&model=anthropic/claude-sonnet-5}}

\end{document}